\documentclass[10pt,twocolumn,letterpaper]{article}
\usepackage[pagenumbers]{cvpr}

\usepackage[T1]{fontenc}
\usepackage{mathtools,bm,multirow,tabularx,array}
\usepackage{tikz}
\usetikzlibrary{arrows.meta,positioning,calc,fit,backgrounds}
\usepackage{microtype}
\usepackage{algorithm}
\usepackage{algpseudocode}
\definecolor{ink}{HTML}{18334A}
\definecolor{teal}{HTML}{197C80}
\definecolor{coral}{HTML}{C66552}
\definecolor{muted}{HTML}{657582}
\definecolor{papergray}{HTML}{F2F5F7}
\definecolor{cvprblue}{HTML}{24608A}
\usepackage[pagebackref,breaklinks,colorlinks,allcolors=cvprblue]{hyperref}
\usepackage[capitalize,noabbrev]{cleveref}

\newcommand{\norm}[1]{\left\lVert#1\right\rVert}

\newcommand{\R}{\mathbb{R}}

\newcolumntype{Y}{>{\raggedright\arraybackslash}X}
\tikzset{box/.style={draw=ink!45,rounded corners=2pt,fill=papergray,align=center,font=\small,inner sep=6pt},flow/.style={-{Stealth[length=4pt]},thick,draw=ink!75}}

\def\confName{CVPR}
\def\confYear{2027}
\title{Seeing Before Naming: Context-Grounded Learning\\of Novel Visual Concepts}

\author{
Gabriel Adineera\\
Texas A\&M University-Commerce\\
USA
\and
Jinoh Kim\thanks{Corresponding author.}\\
Texas A\&M University-Commerce\\
USA
}
\hypersetup{pdfauthor={Gabriel Adineera and Jinoh Kim}}

\begin{document}
\maketitle

\begin{abstract}
Learning an unfamiliar visual category requires more than adapting a classifier: the model must connect its visual evidence to a useful semantic name. A pre-trained vision-language model can retain discriminative image features while providing a weak textual anchor for the target concept. We address this imbalance through context-grounded semantic anchoring. Our framework, SynerNet, learns concept-specific name embeddings, exposes them to compatible linguistic contexts, and conditions their representations on visual evidence. Difficulty-aware processing and adaptive objective weighting coordinate this adaptation while retaining the pre-trained backbone. On four VISTA-Beyond domains, SynerNet reaches a mean 16-shot accuracy of 74.35\%, compared with 72.45\% for few-shot name learning. Across 14 domains in the name-masked transfer setting, it improves mean accuracy from 45.17\% to 46.74\%. Ablations identify concept-specific name embeddings and cross-modal context integration as important components. These results support treating concept acquisition as the construction of visually grounded semantic anchors.
\end{abstract}

\section{Introduction}
\label{sec:intro}
A few photographs can make a category visually recognizable before its name becomes meaningful. Vision-language recognition often converts a category name directly into a classifier weight, obscuring this distinction. A model may retain informative shape, texture, and appearance features while poorly representing the corresponding name. Improving vision alone does not repair that cross-modal association.

\begin{figure}[t]
\centering
\includegraphics[width=\linewidth]{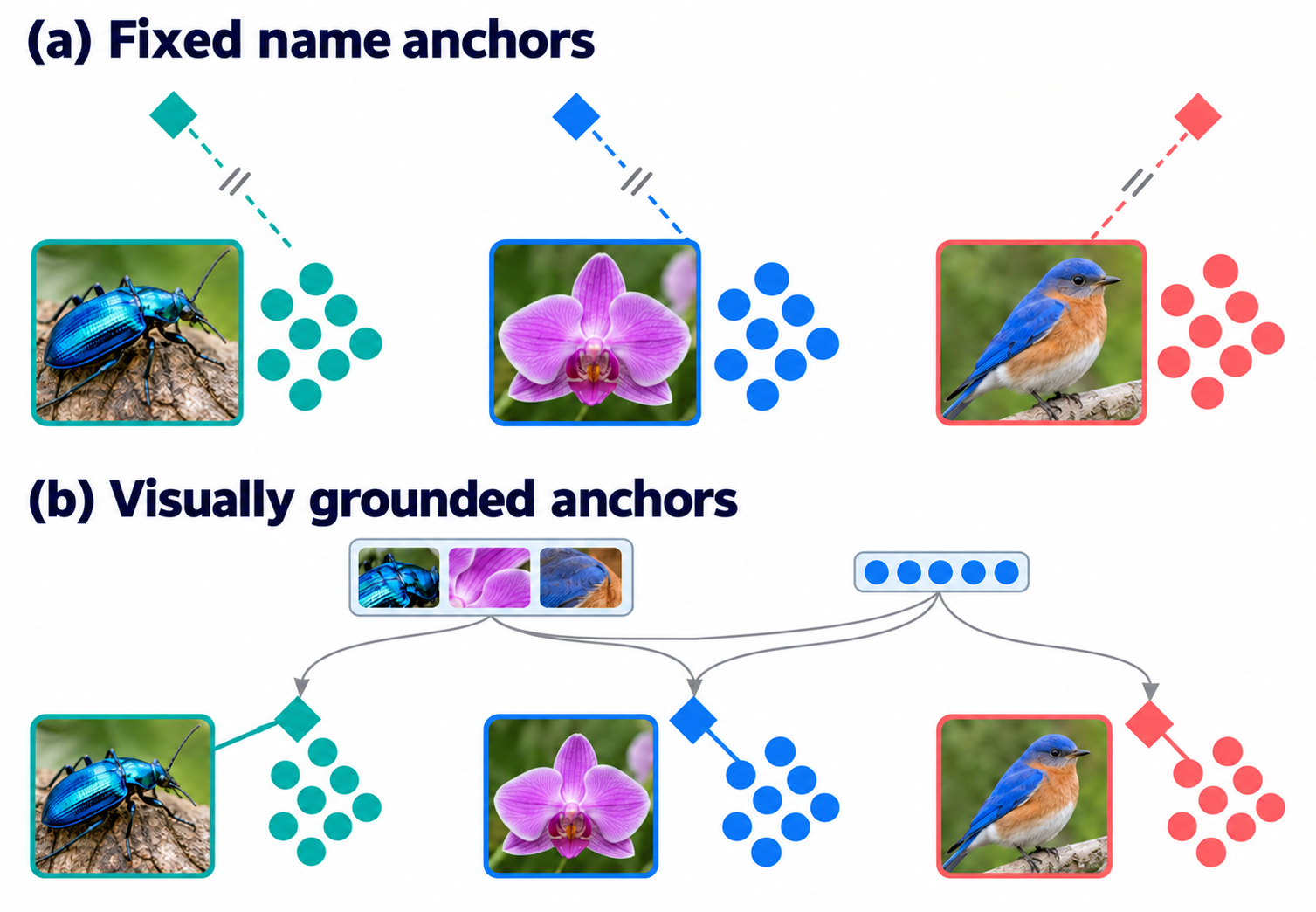}
\caption{\textbf{Visual discrimination need not imply semantic alignment.} Circles denote visual evidence and diamonds denote name anchors. Fixed anchors can remain separated from their corresponding visual groups; name learning and visual context ground the anchors. This is a conceptual schematic with generated example thumbnails.}
\label{fig:concept}
\end{figure}

Contrastive vision-language pre-training, exemplified by CLIP~\cite{clip} and reproducible OpenCLIP models~\cite{openclip}, makes this association useful across a broad range of recognition tasks. Downstream adaptation often learns prompt context~\cite{CoOp,CoCoOp}, transforms features~\cite{clipadapter}, or combines text priors with support-image evidence~\cite{zhang2022tip,huang2024lpp}. These methods provide effective tools for transferring pre-trained knowledge. However, the role of a category name becomes particularly important when the target concept is unfamiliar or poorly aligned with its pre-trained textual representation.

Prior work on learning class names~\cite{NTL} and out-of-pre-training concepts~\cite{chen2025oop} makes this issue explicit through name learning. We build on that motivation and ask a more operational question: \emph{how should a name representation adapt when its most reliable evidence comes from a small collection of images?} A name should preserve the reusable structure of language while being anchored to the visual distinctions that define the current task. Learning the token in isolation leaves the quality of its surrounding context unresolved. Conditioning a text representation on visual evidence without preserving a stable name representation can instead make the semantic anchor too dependent on the current example.

Our central idea is \textbf{context-grounded semantic anchoring}. The concept identity is carried by learnable name embeddings. Compatible linguistic contexts expose that identity under different descriptions. Visual context then adjusts the resulting text representation toward the evidence available in the adaptation task. Together, these operations connect concept identity, description, and visual evidence.

\textbf{SynerNet} combines four interacting modules. The visual perception unit extracts evidence and context; the nominal embedding unit learns concept identity; the linguistic context unit integrates visual and language information; and the coordinator balances alignment and classification. Together, they construct the semantic anchor.

Adaptation difficulty also varies across images and concepts. A single fixed feature-processing rule or loss balance may overemphasize unstable visual evidence or underweight the name representation. SynerNet therefore uses a feature-based difficulty estimate to control processing and learnable weights to balance its objectives. 

The experimental results show complementary effects across settings. In 16-shot recognition, SynerNet improves the four-domain mean over FSNL by 1.90 percentage points, with gains on each evaluated domain. In name-masked transfer, its mean accuracy is 46.74\% across 14 domains, versus 45.17\% for ZSNL. The gain is not uniform: DTD remains a difficult case, and FolkArt favors TransCLIP. The ablations further show that removing the name-learning unit produces the largest mean reduction among the individual module removals.

Our contributions are:
\begin{itemize}
\item A formulation of novel-concept adaptation around context-grounded semantic anchors, connecting concept-specific name learning with compatible linguistic variation and visual conditioning.
\item A coordinated adaptation framework that combines these anchors with difficulty-aware visual processing and adaptive alignment--classification optimization.
\item An evaluation across few-shot recognition, name-masked transfer, and component ablations that identifies both improvements and domain-dependent limitations.
\end{itemize}

Throughout the paper, ``novel'' refers to target concepts in the adaptation task. An assertion that a concept was absent from pre-training requires a separate corpus audit and is not implied by this terminology.

\section{Related Work}
\label{sec:related}
\paragraph{Vision-language representation learning.}
CLIP~\cite{clip} and ALIGN~\cite{ALIGN} learn transferable image-text representations; OpenCLIP~\cite{openclip} and LAION-400M~\cite{schuhmann2021laion400mopendatasetclipfiltered} support reproducible pre-training. BLIP~\cite{li2022blip} and ALBEF~\cite{li2021alignfusevisionlanguage} combine alignment with multimodal objectives, while DenseCLIP~\cite{rao2021denseclip} extends language guidance to dense prediction. We study downstream concept anchors under limited target supervision.

\paragraph{Parameter-efficient and prompt adaptation.}
Adapters~\cite{adapters} and low-rank updates~\cite{LoRA} adapt selected model parameters. Prompt tuning~\cite{promptuning} and visual prompt tuning~\cite{vL-promptuning} instead adapt inputs. CoOp~\cite{CoOp} learns text context, CoCoOp~\cite{CoCoOp} conditions context on images, CasPL~\cite{MTA2} separates generic and task-specific prompts, and DAPT~\cite{Cho_2023_ICCV} considers visual-textual feature distributions. SynerNet adapts concept-bearing tokens and their interaction with visual context.

\paragraph{Feature adaptation and knowledge preservation.}
CLIP-Adapter~\cite{clipadapter} blends residual features, Tip-Adapter~\cite{zhang2022tip} retrieves support evidence, and TaskRes~\cite{yu2023task} learns classifier residuals. LP++~\cite{huang2024lpp} combines prototypes and text embeddings; CLAP~\cite{silva2024closer} studies class-adaptive regularization. WiSE-FT~\cite{wortsman2022robustfinetuningzeroshotmodels} and robust few-shot fine-tuning~\cite{et1} address knowledge preservation under distribution shift. OOD generalization~\cite{ood3} remains distinct from auditing pre-training concept coverage.

\paragraph{Learning names for unfamiliar concepts.}
Parisot et al.~\cite{NTL} learn class-specific word embeddings from images. Chen et al.~\cite{chen2025oop} introduce few-shot and zero-shot name learning for out-of-pre-training concepts. SynerNet adds visual-context integration, linguistic variation, and coordination. TransCLIP~\cite{transclip} provides a complementary transductive comparison. CF-VLM~\cite{Z3} trains on causal counterfactual edits; our context exchange instead aims to preserve concept identity while varying compatible descriptions.

\paragraph{Grounded and compact multimodal representations.}
MM-CoT~\cite{z6} separately probes visual consistency and logical coherence, while top-down semantic refinement~\cite{z13} plans image captions through hierarchical refinement. These tasks require evidence beyond a plausible linguistic output. HTC-VLM~\cite{z7} combines continuous appearance features with discrete semantic anchors for token compression. Video encoders built on image representations~\cite{zhang2026videoencoders} retain frame-specific evidence before temporal refinement. Our anchors serve class discrimination; the experiments do not evaluate reasoning, captioning, compression, or temporal understanding.

\paragraph{Adaptive computation and coordination.}
AdaViT~\cite{meng2021adavitadaptivevisiontransformers} allocates visual computation by input; DrDiff~\cite{z9} schedules experts during text diffusion. KABB~\cite{z1} selects experts, OSC~\cite{z11} adapts inter-agent communication, and MAT-Agent~\cite{z4} coordinates training choices. GAM-Agent~\cite{z2} uses uncertainty to trigger visual-reasoning debate; Agora~\cite{zhang2026doubtsyourselftradingvisual} trades structured visual uncertainty. Failure-driven workflow refinement~\cite{z12} edits workflows using recurring counterexamples. Alongside learning-dynamics-aware fine-tuning~\cite{z10}, these studies illustrate several levels of adaptation.

Markets of Minds~\cite{zhang2026marketsofminds} evaluates expert assistance by expected gain over default continuation, separating repair benefit from regression risk under shared budgets and service capacity. It distinguishes request difficulty from assistance value and the opportunity cost of allocating scarce resources. SynerNet's module-level routing and objective weighting do not model these population-level opportunity costs.

\section{Context-Grounded Semantic Anchoring}
\label{sec:method}

\begin{figure*}[t]
\centering
\includegraphics[width=\textwidth]{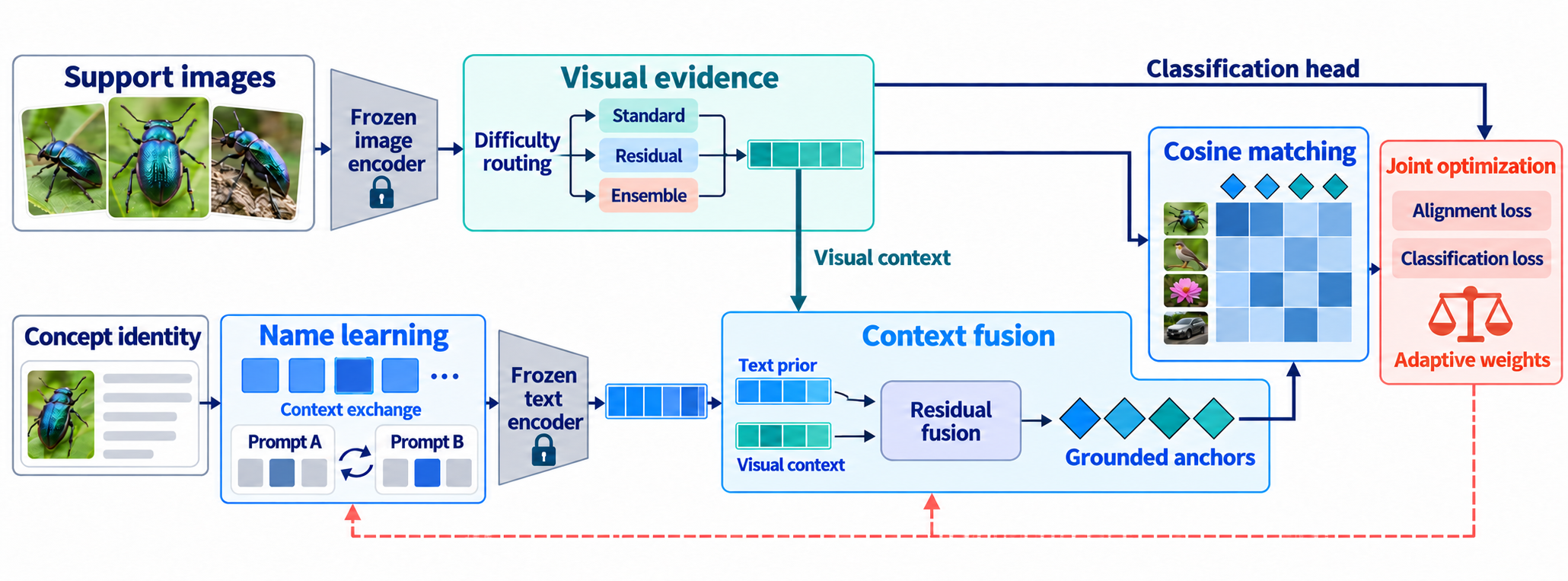}
\caption{\textbf{SynerNet architecture.} Frozen encoders supply image and text features. Name learning and context exchange form concept-specific text representations. Visual context conditions residual fusion to produce semantic anchors for cosine matching. An auxiliary visual classifier and cross-modal alignment provide the joint objectives. Dashed coral arrows denote adaptation feedback. Example thumbnails and the similarity grid are schematic.}
\label{fig:architecture}
\end{figure*}

\subsection{Problem Formulation}
Let $\mathcal Y$ denote the target vocabulary, and let $\mathcal S=\{(x_i,y_i,d_i)\}_{i=1}^{N}$ contain the available adaptation images, class identities, and associated descriptions. In the $K$-shot setting, each target class contributes $K$ labeled images. A pre-trained image encoder $f$ and text encoder $g$ map images and prompts into a shared space. With normalized embeddings $v(x)$ and $t(y)$, the standard classifier is
\begin{equation}
 P(y\mid x)=\frac{\exp(v(x)^\top t(y)/\tau)}
 {\sum_{j\in\mathcal Y}\exp(v(x)^\top t(j)/\tau)}.
 \label{eq:baseline}
\end{equation}
Our objective is to improve the semantic anchor $t(y)$ using limited visual evidence while retaining the reusable representation in the pre-trained backbone.

The framework contains four modules, denoted $\Omega_V$, $\Omega_L$, $\Omega_N$, and $\Omega_C$. Their outputs are visual features, contextualized language features, learned name tokens, and optimization controls, respectively. Information exchange is represented as a typed message
\begin{equation}
 m_{a\rightarrow b}=(a,b,\eta_{ab}),
\end{equation}
where $\eta_{ab}$ contains the feature, context, or control quantity consumed by the receiving module. This notation specifies data flow; the modules operate within a shared differentiable adaptation pipeline.

\subsection{Learning the Concept-Bearing Name}
\label{sec:name}
The nominal embedding unit associates each target concept $y$ with a learnable sequence of tokens
\begin{equation}
 V_y=[e_{y,1},\ldots,e_{y,L_y}],\qquad e_{y,\ell}\in\R^{d_t}.
 \label{eq:tokens}
\end{equation}
A prompt template $\mathcal P_m$ inserts these tokens into a linguistic context. The resulting name-conditioned feature is
\begin{equation}
 t_y^{(m)}=g\!\left(\mathcal P_m(V_y)\right).
 \label{eq:name}
\end{equation}
The tokens carry the target identity, while the template supplies the surrounding description. Learning the tokens allows an unfamiliar category to acquire a representation beyond its original surface string.

\paragraph{Context exchange.}
Given a description $d_i=\mathcal P_i(y_i)$, we retain the concept identity and replace its linguistic context with a context drawn from another adaptation example:
\begin{equation}
 d_i'=\mathcal P_j(y_i),\qquad j\neq i.
 \label{eq:swap}
\end{equation}
The corresponding prompt uses $V_{y_i}$ in both contexts. The purpose is to expose the same name representation under multiple descriptions rather than to create a new visual class. Contexts should remain compatible with the target identity; exchanging a background phrase is different from transferring an attribute that defines another class. This is especially relevant in fine-grained recognition, where a small descriptive change can alter category meaning.

Context exchange reuses the same image-text correspondence; its removal is evaluated in the ablation study.

\subsection{Visual Evidence and Difficulty-Aware Processing}
\label{sec:visual}
The visual unit supplies both the image representation used for recognition and the evidence used to condition the semantic anchor. Its standard representation is $F_V^{\rm std}(x)=f(x)$. The scale-controlled residual representation is
\begin{equation}
 F_V^{\rm res}(x)=\frac{f(x)}{\norm{f(x)}_2}
 +\beta\,\operatorname{sg}(f(x)),
 \label{eq:residual}
\end{equation}
where $\operatorname{sg}$ denotes a stopped-gradient reference. This operation changes the scale entering context fusion; for a fixed image it does not introduce a new feature direction. The distinction matters because cosine normalization removes a pure scale change, whereas the context transformation can still respond to its magnitude.

For a feature summary $\bar v(x)$ produced by the visual unit, the difficulty predictor is
\begin{equation}
 d(x)=\sigma\!\left(W_2\operatorname{ReLU}(W_1\bar v(x)+b_1)+b_2\right).
 \label{eq:difficulty}
\end{equation}
The score controls the selection between standard, residual, and ensemble processing modes. With thresholds $\theta_1<\theta_2$, the routing policy is
\begin{equation}
 \mathcal R(x)=\begin{cases}
 \mathrm{standard},&d(x)<\theta_1,\\
 \mathrm{residual},&\theta_1\leq d(x)<\theta_2,\\
 \mathrm{ensemble},&d(x)\geq\theta_2.
 \end{cases}
 \label{eq:routing}
\end{equation}
The ensemble mode combines available visual processing outputs. Let $F_V(x)$ denote the selected representation. Features are normalized when used for cosine classification, while their context summaries are passed to the linguistic unit. The effect of the difficulty mechanism is evaluated by disabling this routing component.

\subsection{Conditioning the Semantic Anchor on Visual Context}
\label{sec:context}
The linguistic unit receives a name-conditioned text feature and a visual context vector $c$. The latter is an aggregate of the visual features available to the current adaptation batch. A two-layer transformation integrates the modalities:
\begin{equation}
 G_{\rm ctx}(h)=W_4\operatorname{ReLU}(W_3h+b_3)+b_4.
 \label{eq:fusion}
\end{equation}
For a prompt embedding $t_y$, the contextualized anchor is
\begin{equation}
 \widetilde t_y=\lambda t_y+
 (1-\lambda)G_{\rm ctx}([t_y;c]),\qquad 0\leq\lambda\leq1.
 \label{eq:anchor}
\end{equation}
Here $[\cdot;\cdot]$ denotes concatenation. The residual text term preserves the language prior; the transformed term introduces evidence about the visual task. The transformation therefore changes how a learned name is positioned in the shared space without replacing its identity with an image feature.

We normalize $F_V(x)$ and $\widetilde t_y$ before computing their cosine similarity. The resulting classifier retains the form of \cref{eq:baseline}, with the contextualized anchor replacing the fixed category-name feature. The visual context is obtained from model inputs rather than query labels. Because context is aggregated, batching is part of the inference configuration and should remain consistent when comparing methods.

The nonlinear integration and residual mixture distinguish this construction from direct concatenation, which we evaluate as an ablation. The nominal unit supplies the identity; the context unit grounds it in the visual task.

\subsection{Sensitivity of the Grounded Semantic Anchor}
\label{sec:sensitivity}
The residual form gives a direct interpretation of how visual context changes a name representation. Let $t_y\neq0$ be the text prior, $u_y=G_{\rm ctx}([t_y;c])$, and $a_y=\lambda t_y+(1-\lambda)u_y$. Before normalization, its displacement is exactly
\begin{equation}
 a_y-t_y=(1-\lambda)(u_y-t_y).
 \label{eq:displacement}
\end{equation}
Thus the mixing coefficient controls the magnitude of the context correction, while the learned transformation determines its direction. The identity branch remains present throughout this adjustment.

For nonzero $a_y$ and normalized query $v$, normalization obeys
\begin{align}
 \left|v^\top\left(\frac{a_y}{\norm{a_y}_2}
 -\frac{t_y}{\norm{t_y}_2}\right)\right|
 &\leq\frac{2\norm{a_y-t_y}_2}{\norm{t_y}_2}\nonumber\\
 &=\frac{2(1-\lambda)\norm{u_y-t_y}_2}{\norm{t_y}_2}
 \equiv\epsilon_y.
 \label{eq:perturbation}
\end{align}
To see the first inequality, apply Cauchy--Schwarz, expand the normalized-vector difference using the common denominator $\norm{t_y}_2$, and use the reverse triangle inequality for the two feature norms. No assumption about an empirical feature distribution is required.

If the original cosine-score margin between classes $y$ and $j$ is $m_{yj}$, the grounded margin is at least $m_{yj}-\epsilon_y-\epsilon_j$. In particular, a margin larger than this combined perturbation is preserved. This is a sensitivity property of the residual representation, rather than a guarantee of classification correctness. It explains the design tradeoff: a larger context contribution can repair an unsuitable name anchor, while the residual prior limits how strongly that anchor is displaced. The same coefficient should therefore be interpreted together with the scale of the context transformation, not as an independent measure of confidence.

\subsection{Coordinated Optimization}
\label{sec:objective}
The coordinator balances cross-modal alignment and class discrimination. For a batch of $B$ image-text pairs, define $s_{ij}=\langle\widehat F_V(x_i),\widehat{\widetilde t}_j\rangle$, where hats denote normalized features. The symmetric contrastive objective is
\begin{align}
 \mathcal L_{\rm con}&=\tfrac12(\mathcal L_{I\rightarrow T}+\mathcal L_{T\rightarrow I}),\\
 \mathcal L_{I\rightarrow T}&=-\frac1B\sum_i
 \log\frac{\exp(s_{ii}/\tau)}{\sum_j\exp(s_{ij}/\tau)},\\
 \mathcal L_{T\rightarrow I}&=-\frac1B\sum_i
 \log\frac{\exp(s_{ii}/\tau)}{\sum_j\exp(s_{ji}/\tau)}.
 \label{eq:contrastive}
\end{align}
An auxiliary classification head encourages class-discriminative visual features:
\begin{equation}
 \mathcal L_{\rm cls}=-\frac1B\sum_i
 \log\operatorname{softmax}(W_{\rm cls}F_V(x_i))_{y_i}.
 \label{eq:classification}
\end{equation}
The total objective is
\begin{equation}
 \mathcal L=w_{\rm con}\mathcal L_{\rm con}
 +w_{\rm cls}\mathcal L_{\rm cls}.
 \label{eq:objective}
\end{equation}
The coordinator adjusts the two objective coefficients and the temperature, using bounded values to control the optimization scale. The weights determine relative emphasis and are not interpreted as probabilities. The backbone is retained as a pre-trained reference, with adaptation concentrated in concept-specific embeddings and the lightweight coordination and context components.

\begin{algorithm}[t]
\caption{Context-grounded adaptation and recognition}
\label{alg:flow}
\small
\begin{algorithmic}[1]
\For{each adaptation batch}
\State Encode images; select difficulty-dependent features.
\State Build prompts from concept tokens and exchange compatible linguistic contexts.
\State Fuse text features with the visual context.
\State Compute $\mathcal L_{\rm con}$ and $\mathcal L_{\rm cls}$; update adaptive components and coordination parameters.
\EndFor
\State Retain learned tokens and transformations for evaluation.
\State Normalize image and anchor features; predict the highest cosine score over the target vocabulary.
\end{algorithmic}
\end{algorithm}

\paragraph{Adaptation flow.}
Algorithm~\ref{alg:flow} summarizes the computation. Context exchange varies the description of a stable concept, and the coordinator adjusts alignment and classification during adaptation. Recognition then uses the learned anchors under the evaluation configuration.

\section{Experiments}
\label{sec:experiments}

\begin{table*}[t]
\centering\small
\caption{\textbf{Few-shot recognition on four VISTA-Beyond domains.} Top-1 accuracy (\%). Bold indicates the best result within each domain and shot count.}
\label{tab:fewshot}
\begin{tabular}{llccccc}
\toprule
Domain & Method & 1-shot & 2-shot & 4-shot & 8-shot & 16-shot\\
\midrule
\multirow{5}{*}{Insects / spiders} & CoOp & 19.3 & 22.9 & 26.6 & 29.8 & 32.5\\
 & CoCoOp & 16.4 & 17.2 & 17.8 & 19.9 & 22.6\\
 & CLIP-Adapter & 24.3 & 26.8 & 29.5 & 32.8 & 36.0\\
 & FSNL & 31.4 & 35.7 & 39.7 & 41.9 & 43.7\\
 & \textbf{SynerNet} & \textbf{33.8} & \textbf{38.3} & \textbf{41.2} & \textbf{43.5} & \textbf{45.4}\\
\midrule
\multirow{5}{*}{Landmark} & CoOp & 43.8 & 59.9 & 80.1 & 85.8 & 89.4\\
 & CoCoOp & 39.7 & 45.9 & 50.5 & 56.9 & 62.1\\
 & CLIP-Adapter & 57.5 & 69.4 & 85.2 & 88.5 & 91.3\\
 & FSNL & 70.0 & 80.9 & 92.5 & 94.7 & 95.4\\
 & \textbf{SynerNet} & \textbf{72.7} & \textbf{82.8} & \textbf{93.9} & \textbf{96.0} & \textbf{96.7}\\
\midrule
\multirow{5}{*}{Flowers} & CoOp & 35.8 & 54.5 & 81.5 & 87.7 & 90.4\\
 & CoCoOp & 41.6 & 59.4 & 79.1 & 84.3 & 87.8\\
 & CLIP-Adapter & 52.3 & 67.4 & 84.3 & 88.6 & 91.5\\
 & FSNL & 63.4 & 72.7 & 82.9 & 87.3 & 91.0\\
 & \textbf{SynerNet} & \textbf{66.1} & \textbf{75.5} & \textbf{85.8} & \textbf{90.6} & \textbf{93.8}\\
\midrule
\multirow{5}{*}{DTD} & CoOp & 27.5 & 37.0 & 52.4 & 55.8 & 58.4\\
 & CoCoOp & 30.3 & 40.5 & 51.7 & 54.5 & 57.1\\
 & CLIP-Adapter & 39.4 & 45.8 & 52.2 & 57.0 & 59.9\\
 & FSNL & 47.8 & 50.4 & 54.8 & 57.4 & 59.7\\
 & \textbf{SynerNet} & \textbf{49.3} & \textbf{52.6} & \textbf{55.5} & \textbf{59.0} & \textbf{61.5}\\
\bottomrule
\end{tabular}
\end{table*}

\subsection{Experimental Settings}
\paragraph{Evaluation suite.}
We evaluate on VISTA-Beyond: four few-shot domains (Insects / spiders, Landmark, Flowers, and DTD), 14 transfer domains, and five ablation domains. These settings have different target vocabularies and supervision conditions and are analyzed separately.

\paragraph{Few-shot protocol.}
Labeled adaptation examples per class vary over $K\in\{1,2,4,8,16\}$. Adaptation uses OpenCLIP and target image-text pairs, with AdamW and a cosine learning-rate schedule. The configuration considers learning rates between $10^{-5}$ and $10^{-3}$ and three random seeds. We report top-1 accuracy and percentage-point differences.

\paragraph{Name-masked transfer.}
Transfer uses 16 target images per class during adaptation and masks names at inference, testing learned concept representations under restricted name information. This differs from conventional zero-shot recognition without target adaptation. The comparison includes OpenCLIP, TransCLIP, ZSNL, and SynerNet.

\paragraph{Comparators.}
The few-shot baselines are CoOp~\cite{CoOp}, CoCoOp~\cite{CoCoOp}, CLIP-Adapter~\cite{clipadapter}, and FSNL~\cite{chen2025oop}. They respectively adapt prompt context, condition prompts on images, modify features through lightweight adapters, and learn concept-specific name representations. FSNL is especially relevant because it directly addresses the concept name. TransCLIP~\cite{transclip} contributes a transductive transfer comparison; ZSNL~\cite{chen2025oop} is the name-learning comparator in the transfer table.

\subsection{Few-Shot Recognition}
\label{sec:fewshot}
\Cref{tab:fewshot} reports the complete per-domain comparison for the five labeled-data budgets. SynerNet improves on FSNL in all 20 domain--shot combinations. Averaged over the four domains, the gains are 2.33, 2.38, 1.63, 1.95, and 1.90 percentage points for 1, 2, 4, 8, and 16 shots, respectively. 

\paragraph{Fine-grained visual concepts.}
Insects / spiders requires distinguishing categories whose visual differences may be subtle relative to their shared structure. SynerNet reaches 33.8\% with one shot and 45.4\% with 16 shots, compared with 31.4\% and 43.7\% for FSNL. The 2-shot improvement is 2.6 points.

\paragraph{Landmarks.}
The Landmark domain exhibits a strong response to additional labeled examples. SynerNet improves from 72.7\% at one shot to 96.7\% at 16 shots. Its advantage over FSNL is largest at one shot, at 2.7 points, and remains 1.3 points at 16 shots. 

\paragraph{Flowers.}
The Flowers domain shows a sustained advantage across support sizes. SynerNet reaches 85.8\% at four shots and 93.8\% at 16 shots. Its largest gain over FSNL is 3.3 points at eight shots. At 16 shots, CLIP-Adapter is the strongest listed baseline at 91.5\%, giving SynerNet a 2.3-point improvement over the best comparator in that column. 

\paragraph{Texture recognition.}
On DTD, SynerNet reaches 61.5\% at 16 shots, compared with 59.9\% for CLIP-Adapter and 59.7\% for FSNL. The gain over FSNL is smaller at four shots, at 0.7 points, than in the one- or two-shot conditions. Familiar texture words can already provide useful semantic anchors in this domain.

\begin{figure*}[t]
\centering
\includegraphics[width=\linewidth]{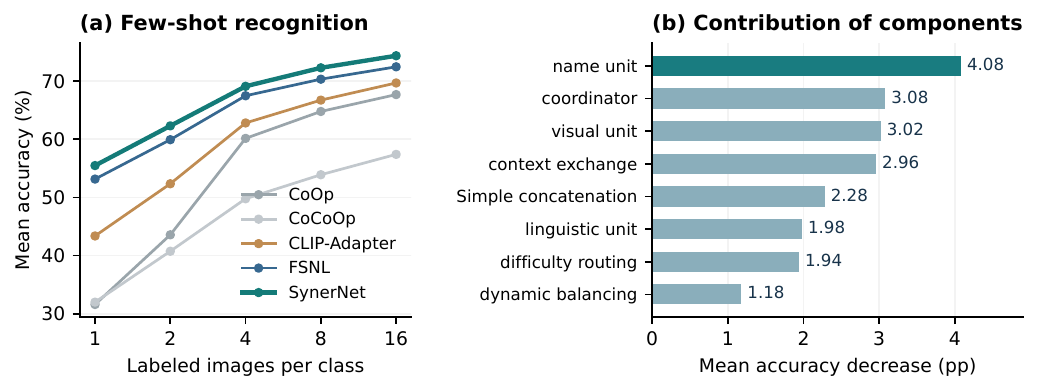}
\caption{\textbf{Accuracy and component contributions.} (a) Macro-average accuracy across the four few-shot domains, computed from Table~\ref{tab:fewshot}. (b) Mean accuracy decrease for each ablation, computed from the five domain scores in Table~\ref{tab:ablation}. The panels summarize different experimental settings and should not be interpreted as sharing the same evaluation population.}
\label{fig:results}
\end{figure*}

\paragraph{Aggregate behavior.}
The macro-average curves in \cref{fig:results} summarize the shot-dependent behavior without pooling images across differently sized domains. At 16 shots, the mean accuracies are 67.68\% for CoOp, 57.40\% for CoCoOp, 69.68\% for CLIP-Adapter, 72.45\% for FSNL, and 74.35\% for SynerNet. The ablations examine the additional roles of context integration and coordination.

\subsection{Name-Masked Transfer Across Domains}
\label{sec:transfer}

\begin{table*}[t]
\centering\small
\caption{\textbf{Component ablations.} Accuracy (\%) in the five-domain ablation setting. The final column is the arithmetic mean decrease from the full model, in percentage points.}
\label{tab:ablation}
\begin{tabular}{lcccccc}
\toprule
Variant & Animals & Architecture & Attire & Food & Pokemon & Mean drop $\downarrow$\\
\midrule
\textbf{Full SynerNet} & \textbf{53.9} & \textbf{89.4} & \textbf{82.0} & \textbf{92.9} & \textbf{79.7} & ---\\
w/o visual unit & 52.1 & 86.3 & 78.3 & 89.8 & 76.3 & 3.02\\
w/o linguistic unit & 52.8 & 87.3 & 79.8 & 90.3 & 77.8 & 1.98\\
w/o name unit & 50.8 & 85.8 & 77.3 & 88.3 & 75.3 & 4.08\\
w/o coordinator & 51.3 & 86.8 & 78.3 & 89.3 & 76.8 & 3.08\\
w/o context exchange & 51.5 & 86.5 & 78.6 & 89.5 & 77.0 & 2.96\\
Simple concatenation & 52.3 & 87.3 & 79.3 & 90.3 & 77.3 & 2.28\\
w/o difficulty routing & 52.5 & 87.6 & 79.8 & 90.7 & 77.6 & 1.94\\
w/o dynamic balancing & 53.3 & 88.3 & 80.8 & 91.3 & 78.3 & 1.18\\
\bottomrule
\end{tabular}
\end{table*}

\begin{table}[t]
\centering\small
\setlength{\tabcolsep}{3pt}
\caption{\textbf{Name-masked transfer across 14 domains.} Top-1 accuracy (\%). Target adaptation uses 16 images per class; names are masked at inference. Bold marks the best score in each row.}
\label{tab:transfer}
\begin{tabularx}{\linewidth}{Ycccc}
\toprule
Domain & OpenCLIP & TransCLIP & ZSNL & Ours\\
\midrule
Animals & 19.4 & 21.7 & 26.4 & \textbf{27.6}\\
Architecture & 22.3 & 24.4 & 39.1 & \textbf{41.3}\\
Attire & 16.8 & 18.8 & 34.5 & \textbf{36.4}\\
FolkArt & 26.7 & \textbf{29.5} & 27.8 & 29.1\\
Food & 8.9 & 10.4 & 29.3 & \textbf{30.7}\\
Insects / spiders & 16.7 & 17.9 & 24.3 & \textbf{26.1}\\
Landmark & 25.8 & 29.6 & 43.9 & \textbf{45.6}\\
Plants & 15.7 & 16.5 & 26.0 & \textbf{30.3}\\
Pokemon & 16.2 & 19.6 & 62.6 & \textbf{64.8}\\
Flowers & 9.9 & 12.7 & 48.3 & \textbf{49.2}\\
Pets & 18.7 & 20.9 & 68.7 & \textbf{71.1}\\
Satellite & 22.3 & 24.7 & 89.5 & \textbf{89.9}\\
DTD & 14.5 & 15.3 & \textbf{44.7} & 44.6\\
UCF101 & 19.9 & 20.6 & 67.3 & \textbf{67.7}\\
\midrule
Mean & 18.13 & 20.19 & 45.17 & \textbf{46.74}\\
\bottomrule
\end{tabularx}
\end{table}

\begin{figure}[t]
\centering
\includegraphics[width=\linewidth]{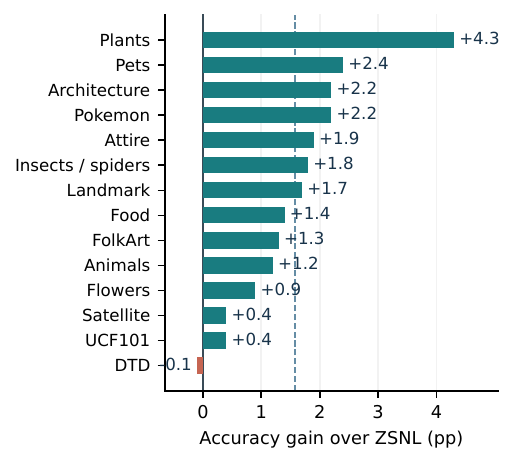}
\caption{\textbf{Domain-level transfer gains.} SynerNet minus ZSNL, computed from Table~\ref{tab:transfer}. The dashed line is the macro-average gain of 1.57 percentage points. Positive gains occur in 13 of the 14 domains.}
\label{fig:transfergains}
\end{figure}

\Cref{tab:transfer} reports the transfer results. SynerNet achieves the best accuracy in 12 of the 14 domains. Its macro-average is 46.74\%, compared with 45.17\% for ZSNL, a gain of 1.57 points. The comparison with the strong name-learning baseline is more informative about the additional contribution of SynerNet than its much larger gap to the unadapted OpenCLIP baseline.

The largest gain over ZSNL occurs on Plants: accuracy increases from 26.0\% to 30.3\%, an improvement of 4.3 points. Other improvements include 2.2 points on Architecture, 2.4 on Pets, and 2.2 on Pokemon. Landmark improves from 43.9\% to 45.6\%. These changes span both natural and constructed visual categories, although they do not imply that every domain benefits equally.

There are two instructive exceptions. On FolkArt, TransCLIP achieves 29.5\%, compared with 29.1\% for SynerNet and 27.8\% for ZSNL. On DTD, SynerNet scores 44.6\%, slightly below ZSNL at 44.7\%. Satellite images also show a small gain over ZSNL, from 89.5\% to 89.9\%. A strong existing semantic anchor or useful unlabeled distribution structure can therefore leave limited room for this adaptation mechanism.

\Cref{fig:transfergains} shows the paired domain-level differences against ZSNL. Thirteen domains improve, while DTD declines by 0.1 points. This differs from the best-of-all-methods count because TransCLIP remains strongest on FolkArt. The distribution of gains also separates the strong Plants improvement from smaller changes in Satellite and UCF101; the macro-average alone obscures that variation.

\subsection{Component Ablations}
\label{sec:ablation}
\Cref{tab:ablation} removes or simplifies individual components in the five-domain setting. The full model averages 79.58\% accuracy over Animals, Architecture, Attire, Food, and Pokemon. Each removal reduces accuracy, but the magnitude varies by module and domain.

\paragraph{Concept-specific name embeddings.}
Removing the nominal embedding unit produces the largest mean reduction, 4.08 points. The drop is 4.7 points on Attire and 4.6 on Food, indicating that the effect is not confined to one category family. This result supports the role of the concept-bearing token representation within the complete model.

\paragraph{Visual and linguistic evidence.}
Removing the visual unit reduces the mean by 3.02 points; removing the linguistic unit reduces it by 1.98 points. The coordinator removal causes a 3.08-point reduction. These results connect the contribution of the name representation to the components that supply evidence, integrate context, and regulate adaptation. They do not by themselves show that the effects are additive or prove a causal interaction between modules.

\paragraph{Context exchange and integration.}
Disabling context exchange reduces the mean by 2.96 points. Replacing nonlinear integration with simple concatenation reduces it by 2.28 points. The difference suggests that exposing a concept to multiple language contexts and learning how those contexts interact with visual evidence are both useful within this architecture. Context exchange alone is not sufficient to explain the performance of the full system.

\paragraph{Adaptive processing and optimization.}
Removing difficulty assessment reduces the mean by 1.94 points, while removing dynamic balancing reduces it by 1.18 points. Their smaller but consistent contributions support their role as stabilizing components around the central anchoring process. The results favor a coordinated adaptation pipeline rather than assigning all improvement to the name embeddings or any single control mechanism.

\subsection{Interpretation}
Few-shot recognition measures sensitivity to labeled support size; name-masked transfer examines restricted inference-time language; ablations identify contributions within the full pipeline. Together, they support learning unfamiliar names through visual evidence and linguistic context while retaining useful pre-trained structure.

\section{Limitations}
\label{sec:limitations}
SynerNet depends on the quality of the pre-trained visual representation. When the encoder does not preserve the distinctions needed for the target task, adjusting a semantic anchor cannot recover the missing visual information. Context exchange also requires care: transferring a class-defining attribute can change the meaning of a description rather than simply diversify its context.

The four-module pipeline introduces more adaptation and coordination than a fixed text classifier. Visual-context aggregation makes batch construction relevant to inference behavior, and difficulty-based routing introduces additional control parameters. The current experiments establish accuracy behavior but do not quantify a latency--accuracy tradeoff. Moreover, the domain-dependent transfer results show that a learned name representation is not universally preferable to strong textual priors or transductive structure. Extending the approach to finer attribute distinctions and more varied naming conventions remains an open direction.

The scope of the evaluation is image recognition. DepthSSC~\cite{z15} aligns depth and spatial evidence for voxel prediction; 3D-Agent's Tri-MARF framework~\cite{Z5} combines views, language, and geometry for 3D annotation. The ActPhysCause challenge~\cite{zhang2026actphyscause} evaluates action-conditioned future videos. The NTIRE super-resolution challenge~\cite{chen2026ntiresr} separates reconstruction fidelity from perceptual quality, while deepfake scaling laws~\cite{wang2025deepfakescaling} concern detector errors across source domains and forgery methods. Extending semantic anchoring to these tasks would require task-specific objectives and evaluation; the present accuracy tables do not establish such capabilities.

\section{Conclusion}
SynerNet adapts novel visual concepts through context-grounded semantic anchoring. It learns concept-bearing name representations, varies their linguistic contexts, and integrates visual evidence into the semantic anchors. Difficulty-aware processing and objective weighting coordinate adaptation. Few-shot, name-masked transfer, and ablation results show gains across domains, with limitations when existing anchors are already strong. The approach emphasizes the learned association between visual evidence and a concept name, alongside the quality of each modality.

\label{maintextend}
\end{document}